# Domain-stratified Training for Cross-organ and Cross-scanner Adenocarcinoma Segmentation in the COSAS 2024 Challenge

Huang Jiayan[1,2], Ji Zheng[1], Kuang Jinbo[1], and Xu Shuoyu[1]

[1] Bio-totem Pte Ltd, Suzhou, China
[2] College of Electronic and Information Engineering, Southwest University, Chongqing, China
Shuoyu.xu@bio-totem.com

**Abstract.** This manuscript presents an image segmentation algorithm developed for the Cross-Organ and Cross-Scanner Adenocarcinoma Segmentation (COSAS 2024) challenge. We adopted an organ-stratified and scanner-stratified approach to train multiple Upernet-based segmentation models and subsequently ensembled the results. Despite the challenges posed by the varying tumor characteristics across different organs and the differing imaging conditions of various scanners, our method achieved a final test score of 0.7643 for Task 1 and 0.8354 for Task 2. These results demonstrate the adaptability and efficacy of our approach across diverse conditions. Our model's ability to generalize across various datasets underscores its potential for real-world applications.

## 1 Introduction

The field of digital pathology has witnessed significant advancements in tumor diagnosis and segmentation, propelled by numerous challenges. However, despite these strides, the effectiveness of current algorithms faces substantial hurdles due to the inherent variability in digital pathology images and tissues. This variability stems from differences in organs, tissue preparation techniques, and image acquisition methods, leading to what is known as domain shift. While specialized methods or network architectures designed for particular types of cancer are common, they often overlook the unique characteristics of different organs and the distinct features produced by varying whole slide scanners.

To tackle these issues, the COSAS challenge offers a chance to assess domain adaptation techniques using an extensive dataset that encompasses a variety of organs and scanners from different manufacturers. Our approach employs organ-stratified cross-validation for Task 1 and scanner-stratified cross-validation for Task 2, training multiple segmentation models aimed at achieving optimal generalization across different organs and scanners. The segmentation results from these models are then ensembled using a voting scheme for Task 1 and probability averaging for Task 2 to produce the final outcomes.

## 2 Method

### 2.1 Data

The dataset provided for Task 1 consists of 60 images from each of three different organs: gastric adenocarcinoma, colorectal adenocarcinoma, and pancreatic ductal adenocarcinoma, totaling 180 images. Each image is derived from a whole slide image (WSI) scanned by the TEKSQRAY SQS-600P scanner, with an average size of 1500 x 1500 pixels.

We employed an organ-stratified three-fold cross-validation technique. In each fold, images from one of the three organ types served as the validation set, while the images from the other two organ types constituted the training set. This methodology ensures that the organ types in each fold of the training and validation sets are completely distinct, enabling us to evaluate the model's generalization capability across different organs and enhancing its robustness and reliability.

For Task 2, we applied a similar scanner-stratified three-fold cross-validation approach. The dataset for this task includes 60 images of invasive breast carcinoma of no special type, captured using three different scanners (TEKSQRAY SQS-600P, KFBIO KF-PRO-400, and 3DHISTECH PANNORAMIC 1000), resulting in a total of 180 images, each approximately 1500 x 1500 pixels. In each fold, we designated all images from one scanner as the validation set, while the images from the other two scanners were used for training. This strategy ensures effective generalization across different scanning devices.

### 2.2 Adenocarcinoma Segmentation

**Model Training**

For our segmentation model, we utilized UperNet [1] with a Visual Attention Network (VAN) [2] serving as the backbone. The VAN incorporates a Large Kernel Attention (LKA) module, enabling self-adaptive and long-range correlations in self-attention. We initiated training with pre-trained weights from ImageNet. Additionally, we augmented the Uper head with an FCN head as an auxiliary network. The outputs of both heads were computed separately using different loss functions and then combined into the total loss.

The original images were approximately 1500x1500 pixels in size. To make full use of the dataset, 512x512 patches were extracted with 50% overlapping across the entire image. This approach allowed us to extract a larger number of patches, effectively increasing the utilization of the dataset and enhancing the model's robustness. A binary mask image of size [512, 512] was also extracted according to each image patch, where 0 and 1 represented the background area (BG) and the cancer area (CA), respectively.

The pixel counts for the background and tumor classes were imbalanced, as some images contained only background with no tumor regions. To address this issue, a data sampling strategy was implemented to balance the number of pixels from both classes. Each original image was divided into 512 x 512 pixel patches, as previously mentioned, and a weight was assigned to each patch based on the ratio between the number of pixels belonging to the two classes. These weights influenced the selection frequency of each patch during training, with patches that had higher weights (indicating more tumor pixels than background pixels) being sampled more frequently. This approach aimed to achieve a balance of pixels from both classes.

Image augmentation techniques were employed to enhance the robustness of the model. These included random rotations within the range of [-45, +45] degrees, random counterclockwise 90° rotations (1, 2, or 3 times), symmetric transformations, Gamma enhancement, contrast enhancement, histogram equalization, solarization, color augmentation in the HSV channel, Gaussian blurring, and random scaling. Each augmentation technique had a 50% probability of being applied, allowing for a diverse mix of transformations that enhanced the model's generalization capabilities.

For the loss function, we employed the sum of Dice Loss and Cross Entropy Loss, incorporating "label smoothing" and "maximal restriction" strategies in the CE loss calculation. During training, we utilized a cosine warm-up learning rate schedule and the Adam optimizer, training for 40 epochs with a batch size of 8. In each epoch, we sampled 17,000 patches using the described strategy. Model performance was evaluated on the validation set after each epoch, and the best model was selected based on the highest average Dice score for both the background and tumor classes.

**Model Inference**

During the image prediction process, the original image was divided into 512 x 512 pixel patches with 50% overlap. A probability heatmap for the tumor class was generated for each patch and subsequently multiplied by a Gaussian kernel. The overlapping areas between adjacent patches were averaged to create the final heatmap. In this final heatmap, pixels with a probability greater than 50% were classified as tumor pixels, while the remaining pixels were assigned to the background class.

**Model Ensemble**

Since we trained three models using three-fold cross-validation for each task, we employed two strategies to ensemble the results from these models. The first method is a hard voting approach. A binary segmentation mask was generated for each model prior to ensemble. For each pixel, the class was assigned based on majority voting from the classification results of the three models.

The second method involved probability averaging. During the creation of the probability heatmap, we averaged the probabilities for each pixel across the three models. The final class for each pixel was determined by selecting the category with the highest probability from the averaged probability heatmap.

For each task, we submitted the results from both ensemble methods during the preliminary test phase and selected the method with the better performance score for the final test phase submission.

## 3 Results

The model performances for both organ-stratified and scanner-stratified three-fold cross-validation are summarized in Table 1, with Dice scores utilized as the primary metrics for evaluation.

Table 1. Three-Fold Cross-Validation Results.

| Task1 | Train: colorectum and pancreas Valid: stomach | Train: colorectum and stomach Valid: pancreas | Train: pancreas and stomach Valid: colorectum | Mean |
|---|---|---|---|---|
| DSC | 0.8200 | 0.8266 | 0.9137 | 0.8534 ± 0.0018 |
| Task2 | Train:3d-1000 and kfbio Valid:teksqray | Train:3d-1000 and teksqray Valid:kfbio | Train:kfbio and teksqray Valid:3d-1000 | Mean |
| DSC | 0.8960 | 0.8979 | 0.9893 | 0.9277 ± 0.0019 |

The results for the two ensemble methods submitted during the preliminary test phase are presented in Table 2. The performance metrics were calculated as the average of the Dice and Jaccard scores (DSC0.5 + JSC0.5) on the Grand Challenge platform.

Table 2. Results obtained in the Preliminary Test Phase.

| Task 1 | Hard voting | Probability Averaging |
|---|---|---|
| Score | 0.7625 | 0.7615 |
| Task 2 | Hard voting | Probability Averaging |
| Score | 0.8722 | 0.8728 |

For the final test phase submission, the hard voting method was selected for Task 1, while probability averaging was chosen for Task 2. In this phase, we achieved scores of 0.7643 for Task 1 and 0.8354 for Task 2, placing us 6th and 2nd in the final ranking, respectively.

## 4 Conclusions

Developing new methods to address domain shifts between different organs or scanners in histopathological images is crucial, particularly for the deployment of models in real clinical settings. In this challenge, we employed an Upernet-based segmentation model, which has been successfully utilized in various pathology-related challenges and in-house projects.

To ensure that the model's generalization ability was thoroughly considered when selecting optimized weights, we implemented a domain-stratified cross-validation approach. The results indicated that our current setup demonstrated better performance in handling variations between different scanners than between different organs. Further improvements could be achieved by incorporating domain adaptation techniques into the segmentation model training process.